\documentclass[sigconf]{acmart}
\AtBeginDocument{%
  \providecommand\BibTeX{{%
    \normalfont B\kern-0.5em{\scshape i\kern-0.25em b}\kern-0.8em\TeX}}}

\usepackage{color}
\usepackage{bm}
\usepackage{multirow}
\newcommand\pjl[1]{\textcolor{black}{#1}}

\newcommand{\wy}[1]{\textcolor{black}{#1}}
\newcommand{\re}[1]{\textcolor{black}{#1}}

\copyrightyear{2023}
\acmYear{2023}
\setcopyright{acmlicensed}\acmConference[MM '23]{Proceedings of the 31st ACM International Conference on Multimedia}{October 29-November 3, 2023}{Ottawa, ON, Canada}
\acmBooktitle{Proceedings of the 31st ACM International Conference on Multimedia (MM '23), October 29-November 3, 2023, Ottawa, ON, Canada}
\acmPrice{15.00}
\acmDOI{10.1145/3581783.3611999}
\acmISBN{979-8-4007-0108-5/23/10}

\begin{document}

\title{Toward High Quality Facial Representation Learning}

% \author{Yue Wang, Jinlong Peng, Jiangning Zhang, Ran Yi, Liang Liu, Yabiao Wang, Chengjie Wang}
% \email{imwangyue, ranyi@sjtu.edu.cn}
% \email{jeromepeng, vtzhang, leoneliu, caseywang, jasoncjwang@tencent.com}

\author{Yue Wang}
\authornote{Equal contribution.}
\orcid{0000-0002-9494-9836}
\affiliation{%
  \institution{Shanghai Jiao Tong University}
  \city{Shanghai}
  \country{China}
}
\email{imwangyue@sjtu.edu.cn}

\author{Jinlong Peng}
\authornotemark[1]
\affiliation{%
  \institution{Youtu Lab, Tencent}
  \city{Shanghai}
  \country{China}
  }
\email{jeromepeng@tencent.com}

\author{Jiangning Zhang}
\affiliation{%
  \institution{Youtu Lab, Tencent}
  \city{Shanghai}
  \country{China}
  }
\email{vtzhang@tencent.com}

\author{Ran Yi}
\authornote{Corresponding author.}
\affiliation{%
  \institution{Shanghai Jiao Tong University}
  \city{Shanghai}
  \country{China}
  }
\email{ranyi@sjtu.edu.cn}

\author{Liang Liu}
\affiliation{%
  \institution{Youtu Lab, Tencent}
  \city{Shanghai}
  \country{China}
}
\email{melpancake@gmail.com}

\author{Yabiao Wang}
\affiliation{%
  \institution{Youtu Lab, Tencent}
  \institution{Zhejiang University}
  \city{Shanghai}
  \country{China}
  }
\email{caseywang@tencent.com}

\author{Chengjie Wang}
\affiliation{%
  \institution{Shanghai Jiao Tong University}
  \institution{Youtu Lab, Tencent}
  \city{Shanghai}
  \country{China}
}
\email{jasoncjwang@tencent.com}

\renewcommand{\shortauthors}{Yue Wang, et al.}

\begin{abstract}
Face analysis tasks have a wide range of applications, but the universal facial representation has only been explored in a few works.
In this paper, we explore high-performance pre-training methods to boost the face analysis tasks such as face alignment and face parsing.
We propose a self-supervised pre-training framework, called \textbf{\it Mask Contrastive Face (MCF)}, with mask image modeling and a contrastive strategy specially adjusted for face domain tasks.
To improve the facial representation quality, we use feature map of a pre-trained visual backbone as a supervision item and use a partially pre-trained decoder for mask image modeling.
To handle the face identity during the pre-training stage, we further use random masks to build contrastive learning pairs.
We conduct the pre-training on the LAION-FACE-cropped dataset, a variants of LAION-FACE 20M, which contains more than 20 million face images from Internet websites.
For efficiency pre-training, we explore our framework pre-training performance on a small part of LAION-FACE-cropped and verify the superiority with different pre-training settings.
Our model pre-trained with the full pre-training dataset outperforms the state-of-the-art methods on multiple downstream tasks.
Our model achieves 0.932 NME$_{diag}$ for AFLW-19 face alignment and 93.96 F1 score for LaPa face parsing.
Code is available at https://github.com/nomewang/MCF.
\end{abstract}







\maketitle

\section{Introduction}

\begin{table}[t]
\begin{center}
\caption{Our training setting and comparison with previous facial representation learning methods. Our method is a pure image-based pre-trained framework.}
\label{tab:motivation}
\begin{tabular}{l|c|c}
\hline
Method & Pretrain dataset & Data format  \\
\hline
Previous & Downstream task dataset & Image\\
FaRL~\cite{FaRL} & LAION-Face & Image + Text  \\
Ours & LAION-Face-cropped \& ImageNet & Image \\
\hline
\end{tabular}
\label{tab:setting}
\end{center}
\end{table}

Facial information is a persistent concern of human society.
With the development of artificial intelligence, a great number of 
deep neural networks for facial applications have been proposed.
Most facial applications focus on a single facial task with supervised learning.
However, the supervised method needs a large number of manually labeled data.
Those labeled data are hard to acquire, and the quality of annotations will significantly influence the model performance. 
Besides, it is difficult for the supervised model to transfer to other domains with different label categories.
Learning a good representation helps promote the facial model performance and solve the above problems.

\pjl{Most existing state-of-the-art facial representation learning methods~\cite{DML-CSR,EAGR,AGR-Net} are} based on supervised learning,
which suffers from expensive data cost and poor generalization ability.
Text supervised methods such as~\cite{FaRL} have also been introduced,
but they need a big amount of text and image data pairs and require huge computing resources during the training process. 
\re{We summarize the above method in Table~\ref{tab:setting}.}
In comparison, self-supervised representation learning is easier to implement, without requirements for labeled training data, and reduces the consumption of computing resources through training strategy design.

In this paper, we propose a novel self-supervised facial representation learning method. We first pre-train backbone network in an unsupervised manner, and then reuse the pre-trained backbone in downstream face analysis tasks.
To learn the face semantic structure, we utilize the mask image modeling strategy to guide the model to learn intrinsic connection of the image patch.
In our model, the input images are first randomly masked partially, and the remaining patches are used to predict information of the masked region, \re{with a partially pre-trained decoder}. E.g., for face pre-training tasks, some face regions are randomly masked, and the goal of the pre-trained model is to reproduce those masked face region with the face patch as hints.
Different from the direct reconstruction methods~\cite{mae,simmim} with pixel loss, we aim to reconstruct the mask regions that have the most similar feature map with the original image.
We argue that the feature map provides additional indication during the pre-training process and further improves the facial representation quality.
The downstream task results show that our training strategy helps improve the model performance on multiple face analysis tasks.

We further expect the pre-trained model to have the ability to automatically distinguish different faces, and project similar faces into closer feature vectors.
Inspired by the contrastive learning methods~\cite{moco,byol}, we suppose each face image belongs to a pseudo-class, which is determined by the face identity.
Based on this idea, we further introduce a face-contrastive learning method.
\wy{Previous work use random resize crop to construct the positive pair, however, we argue that the operation could cause an information loss of the whole face structure.}
Since each face is different from the other, we can build positive pair with different random patch and the negative pair as the different faces with different patch.
This contrastive strategy is well integrated into our mask image modeling framework, and plays a complementary role in better facial representation.
We also provide mathematical analysis and conduct experiments to verify the contrastive strategy.

%缩减1/2
Since the pre-training dataset also plays an important role in self-supervised learning, we further mine the potential of LAION-FACE~\cite{FaRL} dataset.
Although previous work~\cite{FaRL} utilizes both images and texts pair of the LAION-FACE, we argue that the text in the dataset has a weak relation with the corresponding images \re{as illustrate in Figure~\ref{fig:LAION-FACE-cropped}}. 
Since the website data is unstable and could be removed anytime, we reorganize the LAION-FACE dataset and provide LAION-FACE-cropped to facilitate further research.

Overall, we summarize our contributions as follows:
\begin{itemize}

\item We propose a 
self-supervised facial representation learning framework with a feature map based mask image modeling and mask patch based contrastive learning strategy.

\item \wy{We design a partially pre-trained decoder for mask image modeling, which is built on feature map to optimize the model for facial representation learning.}

\item \wy{We propose to use different random masks to construct positive pairs for contrastive learning, which is an important complement for our mask image modeling framework.}

\item \re{We refresh several facial analysis tasks' state-of-the-art results with our pre-trained model.}

\end{itemize}

\section{Related Work}
\begin{figure*}
\centering
\includegraphics[width=2\columnwidth]{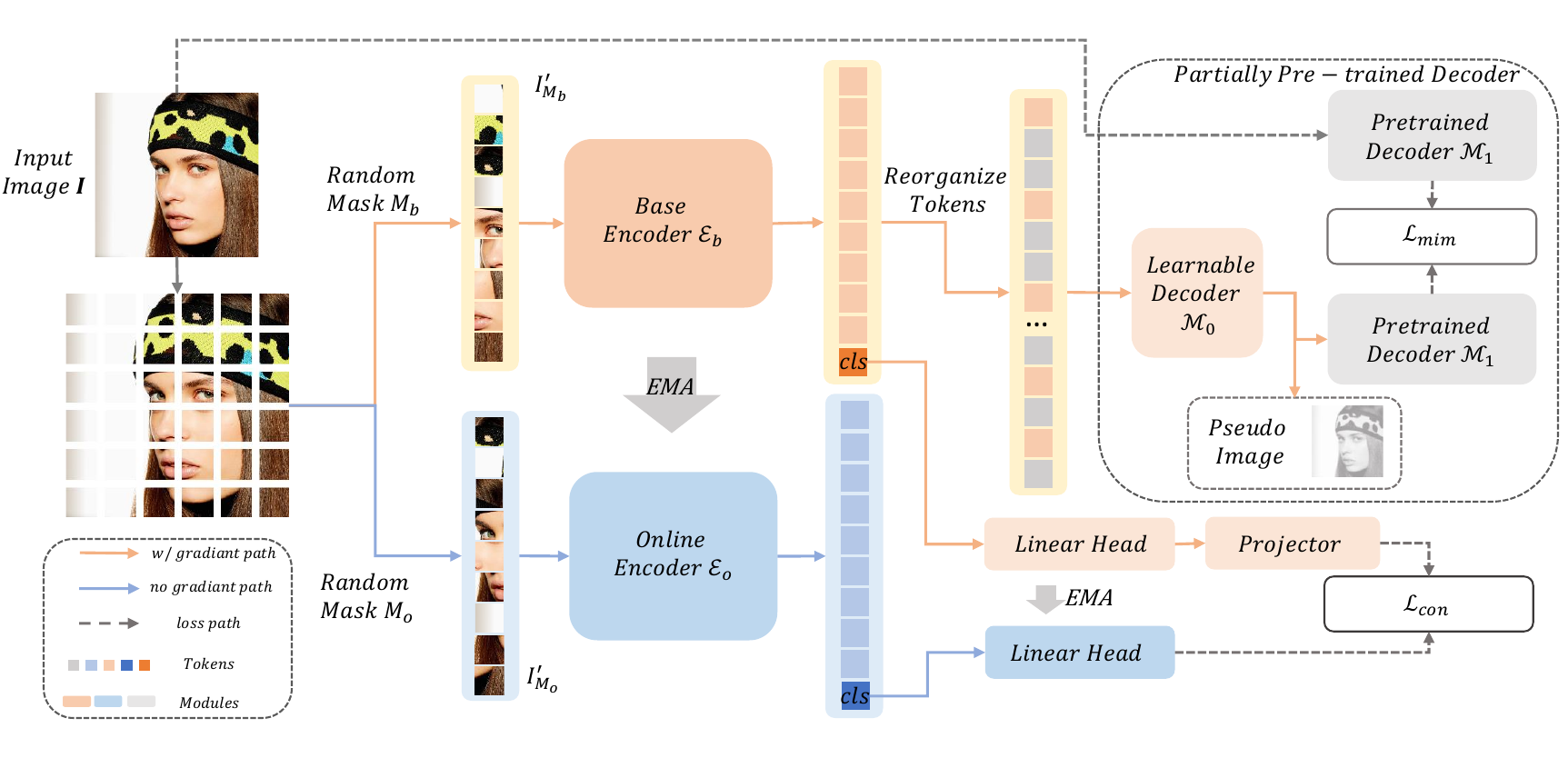}
\caption{The framework of \textit{Mask Contrastive Face (MCF)}. We combine mask image modeling and contrastive strategy for high-quality facial representation learning. Given an input image $I$, we get two masked image patch sequences with different random masks, and input them separately into base encoder $\mathcal{E}_b$ and online encoder $\mathcal{E}_o$. The output patch token of $\mathcal{E}_b$ will be used for partially learnable decoder $\mathcal{M}_0 \circ \mathcal{M}_1$ to predict the feature map of $I$. Furthermore, we utilize the class token ($cls$) of $\mathcal{E}_b$ and $\mathcal{E}_o$ to calculate contrastive loss for enhancing facial representation. (Source image comes from CelebAMask-HQ~\cite{celebamask}.)}
\label{fig:pipeline}
\end{figure*}

\subsection{Facial Representation Learning}

Previous works~\cite{pretrain,preface,FaRL} demonstrate that the pre-train strategy reduces the overfitting and \pjl{improves} model performance in the face analysis area. 
Although some general representation learning methods have achieved good performance, the method designed for face images still has better results than the one training on the general dataset~\cite{FaRL}.
Supervised representation learning methods~\cite{DML-CSR,EAGR,AGR-Net,zhang2020freenet,xu2022designing} need amounts of manually labeled data and suffer from overfitting problems with a large number of model parameters.
Recently some tasks~\cite{CLIP,FaRL} use Web text-image pairs as pre-train dataset, \pjl{which achieve} amazing representation learning results.
We further utilize the Web image dataset, and
get higher performance on downstream tasks with our pre-train framework. 

\subsection{Mask Image Modeling}

Mask Image Modeling (MIM) is a kind of visual self-supervised learning method that raises wide attention recently.
The total manner of MIM is: given a masked image or token as input, the model outputs a full image, and learns a good feature representation for downstream tasks during this process.
MST~\cite{MST} and iBoT~\cite{ibot} propose mask token strategy when designing a knowledge distillation network.
BEiT~\cite{beit} and PeCo~\cite{peco} convert images to visual tokens and train the model in the same manner as BERT~\cite{bert} which predicts the words token for \pjl{language} task pre-train.  
SimMIM~\cite{simmim} and MAE~\cite{mae} propose a simple auto-encoder structure for model pre-train and achieve impressive results with less computation resource consumption than the previous method.
MaskFeat~\cite{maskfeat} predicts the feature map of the input image and study some different types of feature performance.
The above works almost concentrate on general pre-train (e.g. ImageNet~\cite{imagenet}) and have few discussions when it comes to a certain domain (e.g. face, animal).
We concentrate on the face domain, %which has fewer variations between images, and 
where the downstream tasks focus on the semantic information of the face region.

\subsection{Contrastive Self-Supervised Learning}

Contrastive learning is proven as another efficient self-supervised learning strategy, 
which aims to minimize the output feature distance between similar samples.
The mainstream works are based on InfoNCE~\cite{moco} loss and are popular with both the convolution-based model~\cite{moco,byol} and self-attention-based model~\cite{mocov3,dino}.
Recently iBOT~\cite{ibot} use a contrastive strategy to train online tokenizers for mask image modeling.
The models pre-trained with contrastive learning methods~\cite{dino,ibot} can automatically generate a coarse unsupervised semantic segmentation map.  
% Our method also combines contrastive strategy and mask image modeling, 
Different from previous methods, we use the two items as \re{reciprocal} complementarities to each other, and improve the quality of facial representation.

\section{Method}

\subsection{Overview}
\label{sec:overview}

We aim to train a self-supervised facial representation learning model with mask image modeling 
and contrastive learning strategy, called \textit{Mask Contrastive Face (MCF)}.
The framework is illustrated in Figure~\ref{fig:pipeline}. 
Our framework contains two encoders: base encoder $\mathcal{E}_b$ and online encoder $\mathcal{E}_o$, where the online encoder is designed as a self-distillation teacher for contrastive learning.
A partially pre-trained decoder ($\mathcal{M}_0 \circ \mathcal{M}_1$) is designed for mask image modeling, and predicts the feature map of the original image $I$.

For the base encoder, we design a partially pre-trained decoder for mask image modeling.
Different from other image reconstruction mask image modeling~\cite{mae,simmim}, instead of using pixel loss as a training objective function, the training goal of our method is to fit the output image feature map with the input image feature map.
Our method is closer to MaskFeat~\cite{maskfeat}, which uses a decoder to directly predicts the image feature map instead of the original image as the final output.
We argue that the task of predicting feature map is too complex for a single decoder, and the model will pay more attention to fitting the target feature map instead of learning a general image representation.
We set a light decoder first to generate a pseudo image, which is then sent into a pre-trained visual backbone to generate the target feature map $F_{t}$.
The original image is also sent into the pre-trained visual backbone to get the ground truth feature map $F_{o}$.
Then the mask image modeling training objective function is used to minimize the distance of the above two feature maps.

The online encoder is designed for contrastive learning, based on the assumption that each face could be distinguished by a self-supervised model without any additional label,
and similar faces should have a closer feature distance.
Different from the previous contrastive learning methods~\cite{moco,mocov3,dino}, we use random masks to construct the positive pair and negative pair. 
\wy{Without random cropping, our method keeps the whole face structure and the model can learn more face identity information. }
We set the positive sample as the same face image masked by a different random patch, and the negative sample as a different face image masked by random patch. 

The mask image modeling part and contrastive learning part work together to get high-quality facial representation.
The backbone pretrained by our learning strategy is able to be used in the downstream tasks, and can achieve high performance.% with transfer training. 

\subsection{Analysis of Pre-training Process}
\label{sec:math}

To enhance representation learning, we aim to find an efficient optimization method for the self-supervised pre-training process.

Give a target encoder $E$ and an image patch $x$, during backward optimization, the gradient is related to the partial derivative $\frac{\partial{E}}{\partial{x}}$. And if we add more modules during training, the partial derivative will be more complex.
For mask image modeling, the model consists of a target encoder $E$ and an auxiliary $n$-layer decoder $T_0 =t_1\circ ...\circ t_n$, and the whole model can be denoted as $f = T_0\circ E $, then the partial derivative can be formulated as:
\begin{equation}
    \frac{\partial{f}}{\partial{x}} = \frac{\mathrm{d}T_0}{\mathrm{d}E}\frac{\partial{E}}{\partial{x}} = \frac{\mathrm{d}t_1}{\mathrm{d}t'_1}\cdot...\cdot\frac{\mathrm{d}t_n}{\mathrm{d}E}\cdot\frac{\partial{E}}{\partial{x}},
\end{equation}
where $t'_1 = t_2 \circ ...\circ t_n$. As $n$ becomes larger, the $\frac{\partial{E}}{\partial{x}}$ becomes more uncertain, which explains why previous self-supervised pre-training methods tend to use a lightweight decoder.

With a heavier decoder $T_1=t_1\circ ...\circ t_m, m>n$, the partial derivative is:
\begin{equation}
    \frac{\partial{f}}{\partial{x}} = \frac{\mathrm{d}T_1}{\mathrm{d}E}\frac{\partial{E}}{\partial{x}} = \frac{\mathrm{d}t_1}{\mathrm{d}t'_1}\cdot...\cdot\frac{\mathrm{d}t_n}{\mathrm{d}t'_n}\cdot...\cdot\frac{\mathrm{d}t_m}{\mathrm{d}E}\cdot\frac{\partial{E}}{\partial{x}},
\end{equation}
where the additional part $\frac{\mathrm{d}t_n}{\mathrm{d}t'_n}\cdot...\cdot\frac{\mathrm{d}t_m}{\mathrm{d}E}$ produces more degree of freedom during training.
However, if we fix the additional part and assign pre-trained parameters to this additional part, the optimization process would receive some useful indication, and the partial derivative can be formulated as:
\begin{equation} 
\frac{\partial{f}}{\partial{x}} = \frac{\mathrm{d}T_1}{\mathrm{d}E}\frac{\partial{E}}{\partial{x}} = \frac{\mathrm{d}t_1}{\mathrm{d}t'_1}\cdot...\cdot\frac{\mathrm{d}t_n}{\mathrm{d}n'}\cdot\alpha\cdot\frac{\partial{E}}{\partial{x}} 
= \alpha \cdot \frac{\mathrm{d}T_0}{\mathrm{d}E}\frac{\partial{E}}{\partial{x}},
\end{equation}
where $\alpha$ is the additional part with fixed pre-trained parameters.

Then the decoder can also be extended to a mixed one, which can be denoted as $T = M + C$. If $M$ is a partial pre-trained module and C is a much smaller scratch network, we can get the partial derivative as:
\begin{equation}
    \frac{\partial{f}}{\partial{x}} = \alpha \cdot \frac{\mathrm{d}M_0}{\mathrm{d}E}\frac{\partial{E}}{\partial{x}} + \frac{\mathrm{d}C}{\mathrm{d}E}\frac{\partial{E}}{\partial{x}}=(\alpha \cdot a + b)\frac{\partial{E}}{\partial{x}},
    \label{eq:linear}
\end{equation}
where we set $a=\frac{\mathrm{d}M_0}{\mathrm{d}E}, b= \frac{\mathrm{d}C}{\mathrm{d}E}$. The auxiliary optimization becomes similar to linear programming task, which simplifies the encoder optimization and promotes the model performance.
In our framework, the decoder of $M$ is implemented as our mask image modeling with a partially pre-trained decoder and the $C$ is a head optimized by contrastive strategy. 

\subsection{Partially Pre-trained Decoder for Mask Image Modeling}
\label{sec:mim}

As discussed in Sec~\ref{sec:math}, we design a partially pre-trained decoder for mask image modeling as the scale item in the linear-programming-like optimization process in Eq.(\ref{eq:linear}).
The goal of our mask image modeling is to predict the feature map $F$ of the original input image $I$, from the image $I_{M_b}$ masked by a random mask $M_b$.
We expect the model to maximize $\log q_{\theta}(F|I_{M_b})$, where $q_{\theta}$ is predicted feature distribution with learnable parameters $\theta$.

In detail, with a random mask $M_b$, we remove the masked patch and get a sequence of unmasked image patches $I'_{M_b}$ \wy{(as shown in Figure~\ref{fig:pipeline})}.
We denote the base encoder as $\mathcal{E}_b$ and the base encoder output (a sequence of latent feature vectors) as $F_b$.
For the decoder, we separately denote the learnable part as $M_0$ and the fixed part as $M_1$.
The $M_0$ outputs a pseudo image, and the $M_1$ predicts the feature map of input face image $I$.
Instead of minimizing the pixel loss as previous methods~\cite{mae,simmim}, we minimize the feature map difference between the original face image $I$ and the pseudo image outputted by $M_0$.
The loss function can be formulated as:
\begin{equation}
    \mathcal{L}_{mim}=\Vert M_1 \circ M_0(\mathcal{E}_b(I'_{M_b})) - M_1(I)\Vert_2.
\end{equation}

During the backward stage, we calculate the gradient with the indication of the pre-trained parameters in $M_0$. 
We use a network pre-trained on ImageNet1k as $M_0$ for general visual knowledge.

\subsection{Random Mask for Contrastive Learning}
To encode face images into a disentangle latent space, we design a contrastive learning branch as the bias item in Eq.(\ref{eq:linear}).
Given two different masked face images, we expect the model to determine if the images come from the same face.
Each face can be seen as a single class, and face contrastive learning can be implemented by cross-entropy loss and self-distillation.

\wy{Previous works use random cropping to build the positive pair belonging to the same class. But that operation could be harmful for the model to learn face identity.}  
For the input face image $I$, with two different masks $M_b$ and $M_o$, we can get two masked face image $I_{M_b}$ and $I_{M_o}$ as positive pair, we expect that the model can recognize they come from the same face image.
We use the class token of the Vision Transformer~\cite{ViT, zhang2021analogous, zhang2022eatformer, zhang2023rethinking} as the identifying label and a teacher-student framework to get the predictive categorical distributions.

The encoder $\mathcal{E}_b$ used in Sec~\ref{sec:mim} is reused as a student network, then we design a teacher network $\mathcal{E}_o$ which has the same structure as $\mathcal{E}_b$  and the parameters of $\mathcal{E}_o$ come from past iterations of $\mathcal{E}_b$.
And the goal of contrastive learning is to minimize ${P}_o(I_{M_b}) \log {P}_b(I_{M_b})$, where ${P}_o$ and $P_b$ are the posterior distributions with $\mathcal{E}_o$ and  $\mathcal{E}_b$ respectively.
\wy{ For implementation, we use a linear head and a projector to project the output feature of $\mathcal{E}_b$ to query vector $q$, and only use a linear head to project the output feature of $\mathcal{E}_o$ to key vectors $k$.}
The final contrastive loss is formulated as:
\begin{equation}
\mathcal{L}_{con} = -log\frac{\exp(q \cdot k^T/\tau)}{\sum_{i=1}^{N_b} \exp(q \cdot k_i/\tau) },
\end{equation}
where $N_b$ is batch size. The parameters of %the base encoder 
$\mathcal{E}_b$ are Exponentially Moving Averaged (EMA)~\cite{moco} to the parameters of $\mathcal{E}_o$.

\subsection{Noised Website Face Data}
\label{sec:laion-face-cropped}

\begin{figure}
\centering
\includegraphics[width=0.9\columnwidth]{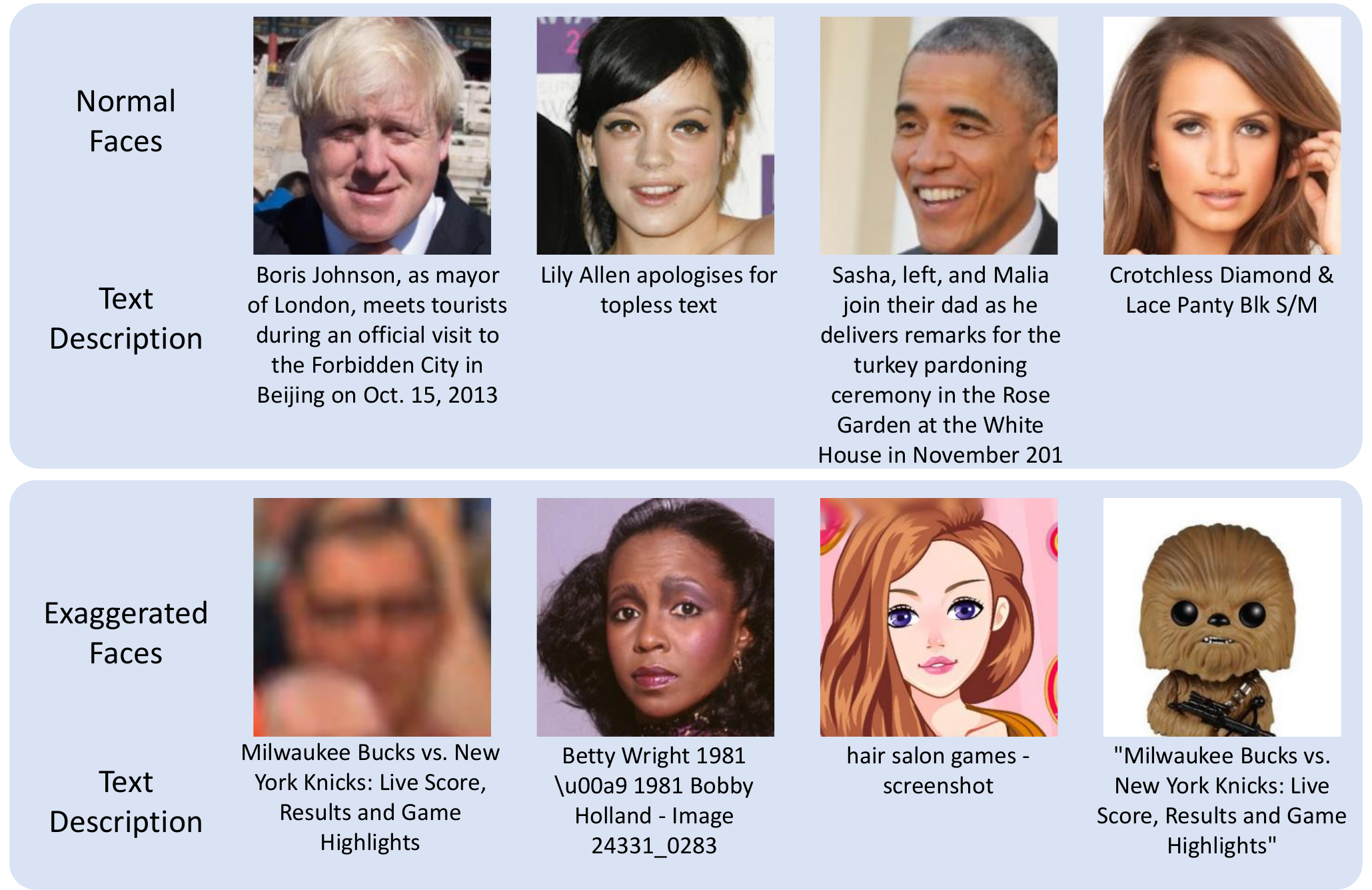}
% \vspace{-5pt}
\caption{Samples of LAION-FACE-cropped. We crop the face region and align faces to the average face. It contains the real human face and a small part of the exaggerated face (e.g. blurred face, artistic face, etc). We keep the text description of the original image, and there is a weak semantic relation in most image-text pairs. (Source images come from  LAION-FACE~\cite{FaRL,laion}.)}
% \vspace{-5pt}
\label{fig:LAION-FACE-cropped}
\end{figure}

The pre-training dataset plays an important role in facial representation learning.
We utilize the LAION-FACE~\cite{FaRL} dataset for our pre-training process, and modify the face dataset for better self-supervised pre-training.
The original LAION-FACE is a website dataset containing 20 million image-text pairs, and the FaRL~\cite{FaRL} utilizes those pairs to learn facial representation in a visual-linguistic manner.
But the LAION-FACE dataset has some shortages:
(1) The websites data is unstable, some datasets could be unreachable as time going;
(2) Some text description is for the original image and the inherent connection with the face is not intensive enough, as shown in Figure~\ref{fig:LAION-FACE-cropped}.
To solve problem 1, we download the LAION-FACE dataset as completely as possible and reorganized the LAION-FACE dataset by every single face.
In detail, we collected 65\% data of the LAION-FACE dataset, which is determined by our network quality and website accessibility.
We utilize a face detection network to detect the face region of our downloaded LAION-FACE images and report 5 face landmarks for each face.
Then we crop detected faces out and align them to the average face~\cite{stylegan}.
Finally, we resize the cropped face image to 256$\times$256 resolution and store the resized image and text description as LAION-FACE-cropped dataset.
For problem 2, during our pre-training process, we only utilize the face images of LAION-FACE-cropped. Our experiment shows that we can still get impressive results without the text part.

\section{Experiments}

\subsection{Pre-train Setup}

We modify LAION-FACE~\cite{FaRL} dataset and construct LAION-FACE-cropped to pre-train the backbone model.
LAION-FACE-cropped contains about 20 million face images processed from images in LAION-FACE 20M dataset as described in Sec~\ref{sec:laion-face-cropped}.
We set a light setting that uses about 10\% of LAION-FACE-cropped for method evaluation and a full setting for high-quality facial representation.
We use ViT-b/16 as our visual backbone whose input image patch size is $16\times16$ resolution, indeed there will be $14 \times 14$ patches for $224 \times 224$ input resolution in ViT-b/16. 
For mask strategy, we follow the setting of MAE~\cite{mae}, and only the unmasked token is used during pre-train.
For the decoder, $\mathcal{M}_0$ is a 2-layer transformer, and $\mathcal{M}_1$ is a transformer pre-trained with ImageNet dataset~\cite{imagenet}.
We set the mask ratio as 0.75, which means only 50 visual tokens is sent to the Encoder ViT.
The batch size is 4,096, and the weight decay is 0.05.
We train 16 epochs on LAION-FACE-cropped datasets and 1 epoch for linear warmup.
We use 8 Nvidia Tesla V100 and the total pre-train time is about 10 hours for the light setting and about 4 days for the full set.

\subsection{Implementation Details}
\label{sec:imple}

In this section, we provide more implementation details of the Mask Contrastive Face.
For the downstream task, we use the pre-trained encoder $\mathcal{E}_b$, which is a ViT-b/16 backbone.
We use the outputs of the \{2, 4, 8, 12\}-th layers of $\mathcal{E}_b$ to build a feature pyramid network, and then utilize UperNet to generate face alignment heat maps or face parsing segmentation maps as the same as~\cite{FaRL}. 
The decoder $\mathcal{M}_0$ contains two transformer blocks and outputs 512 dimension feature vectors.
The encoder output does not contain the masked patch information, hence we add original image position information to the encoder output and use a mask token for the missed patch.
A 1-layer linear projector converts the decoder transformer blocks outputs to pseudo images.
The decoder $\mathcal{M}_1$ is another ViT-b/16 network, pre-trained with ImageNet 1K dataset.
For a more explainable feature map, we use contrastive learning to pre-train $\mathcal{M}_1$ and follow the training details of MoCo-v3~\cite{mocov3}. 
For the contrastive part, the online encoder $\mathcal{E}_o$ has the same structure as $\mathcal{E}_b$.
The linear head contains 2 linear layers and projects the class token to a 256 dimension vector.
The projector also contains 2 linear layers, but is a 256-dimension to 256-dimension transformation.

We pre-train the $\mathcal{E}_b$ on 8 Nvidia Tesla V100 GPU.
The total batch size is 512, and the batch size for each GPU is 64.
We accumulate 8 iterations of gradient together for a single optimization, which means the equivalent batch size is 4,096.
The learning rate is warmed up to 0.016 with 1 epoch, and cosine decayed to 0 with the last 15 epochs.
Finally, the weight decay is set as 0.05.

\subsection{Comparison with State-of-The-Art Face Analysis Methods}

\subsubsection{Face Alignment}

\begin{table}[t]
\begin{center}
\caption{Comparison with the state-of-the-art AFLW-19 face alignment methods. Bold font is the best value and underline is the second best.}
\label{tab:AFLW}
\setlength\tabcolsep{3pt}
\renewcommand\arraystretch{1}
\footnotesize
\begin{tabular}{l|cc|cc}
\hline
\multirow{2}*{Method} & \multicolumn{2}{c|}{NME$_{diag}$ $\downarrow$}  & NME$_{box}$  $\downarrow$ & AUC$^{7}_{box}$ $\uparrow$\\
~ & Full & Front & Full & Full\\
\hline
CFSS & 3.92 & 2.68 & - & - \\
CCL & 2.72 & 2.17 & - & - \\
DAC-CSR & 2.27 & 1.81 & - & - \\
LLL & 1.97 & - & - & - \\
SAN & 1.91 & 1.85 & 4.04 & 54.0 \\
DSRN & 1.86 & - & - & - \\
LAB & 1.25 & 1.14 & - & - \\
HR-Net & 1.57 & 1.46 & - & - \\
Wing & - & - & 2.80 & 60.3 \\
Bulat et al. & 1.54 & - & - & - \\
KDN & - & - & 2.80 & 60.3 \\
LUVLi & 1.39 & 1.19 & 2.28 & 68.8\\
% FaRL* & 0.986 & 0.836 & 1.39 & 80.5\\
\hline
Scratch & 1.047 & 0.884 & 1.481 & 79.3 \\
FaRL & 0.969 & 0.836 & 1.371 & 80.8 \\
Ours $_{0.1}$ & 0.968 & 0.827 & 1.368 & 80.8 \\
Ours & 0.950 & 0.820 & 1.344 & 81.2 \\
\hline
FaRL$^{448}$ & \underline{0.943} & \underline{0.821} & \underline{1.334} & \underline{81.3}\\
Ours$_{0.1}^{448}$ & 0.951 & 0.825 & 1.345 & 81.2 \\
Ours$^{448}$ & {\bf 0.932} & {\bf 0.810} & {\bf 1.318} & {\bf 81.5}\\
\hline
\end{tabular}
\end{center}
\end{table}

\begin{table}[t]
\begin{center}
\caption{Comparison with the state-of-the-art WFLW face alignment methods. Our method outperforms previous methods in both 224$\times$224 and 448$\times$448 resolution.}
\label{tab:WFLW}
\setlength\tabcolsep{1.5pt}
\renewcommand\arraystretch{1}
\footnotesize
\begin{tabular}{l|c|cccccc|cc}
\hline
\multirow{2}*{Method} & \multicolumn{7}{c|}{NME$_{ioc}$ $\downarrow$} &FR$^{10}\downarrow$ & AUC$^{10}\uparrow$ \\
~ & Full & Pose & Expr. & Illum. & MakeUp & Occl. & Blur & \multicolumn{2}{c}{Full}\\
\hline
ESR & 11.13 & 25.88 & 11.47 & 10.49 & 11.05 & 13.75 & 12.20 & 35.24 & 27.74 \\
SDM & 10.29 & 24.10 & 11.45 & 9.32 & 9.38 & 13.03 & 11.28 & 29.40 & 30.02 \\
CFSS & 9.07 & 21.36 & 10.09 & 8.30 & 8.74 & 11.76 & 9.96 & 20.56 & 36.59 \\
DVLN & 6.08 & 11.54 & 6.78 & 5.73 & 5.98 & 7.33 & 6.88 & 10.84 & 45.51 \\
LAB & 5.27 & 10.24 &  5.51 & 5.23 & 5.15 & 6.79 & 6.12 & 7.56 & 53.23 \\
Wing & 5.11 & 8.75 & 5.36 & 4.93 & 5.41 & 6.37 & 5.81 & 6.00 & 55.04 \\
DeCaFa & 4.62 & 8.11 & 4.65 & 4.41 & 4.63  & 5.74 & 5.38 & 4.84 & 56.30 \\
Bulat et al & 4.57 & - & - & - & - & - & - & - & -  \\
AWing & 4.36 & 7.38 & 4.58 & 4.32 & 4.27 & 5.19 & 4.96 & 2.84 & 57.19 \\
LUVLi & 4.37 & - & - & - & - & - & - & {\underline{2.84}} & 57.19 \\
ADNet & 4.14 &  6.96 &  4.38 &  4.09 &  4.05 &  5.06 &  4.79 & 2.72 & 60.22 \\
\hline
Scratch & 4.80 & 8.78 & 5.09 & 4.74 & 4.99 & 6.01 & 5.35 & 5.72 & 54.54 \\
FaRL & 4.03 & 6.81 & 4.32 & 3.92 & 3.87 & 4.70 & 4.54 & 1.76 & 60.23 \\
Ours$_{0.1}$ & 4.16 & 7.15 & 4.40 & 4.05 & 4.09 & 4.99 & 4.66 & 2.56 & 59.19 \\
Ours & \underline{3.96} & \underline{6.61} & \underline{4.13} & \underline{3.84} & \underline{3.80} & \underline{4.68} & \underline{4.43} & \bf{1.40} & 60.90 \\
\hline
FaRL$^{448}$ & \underline{3.96} & 6.91 & 4.21 & 3.97 & 3.80 & 4.71 & 4.57 & 1.76 &  61.16 \\
Ours$_{0.1}^{448}$ & 4.13 & 7.18 & 4.31 & 4.09 & 4.07 &  5.00 & 4.72 & 2.44 & 60.18  \\
Ours$^{448}$ & \bf{3.90} & \bf{6.55} & \bf{4.07} & \bf{3.82} & \bf{3.72} & \bf{4.61} & \bf{4.42} & \underline{1.60} & \bf{61.44} \\
\hline
\end{tabular}
\end{center}
\end{table}

\begin{table}[t]
\begin{center}
\caption{Comparison with the state-of-the-art iBUG-300 face alignment methods. Our method outperforms the previous methods on both common and full categories.}
\label{tab:300W}
\setlength\tabcolsep{3pt}
\renewcommand\arraystretch{1}
\footnotesize
\begin{tabular}{l|cc|c}
\hline
\multirow{2}*{Method} & \multicolumn{3}{c}{NME$_{ioc}$ $\downarrow$}\\
~ & Common & Challenge & Full\\
\hline
SAN & 3.34 & 6.60 & 3.98 \\
AVS & 3.21 & 6.49 & 3.86 \\
DAN & 3.19 & 5.24 & 3.59 \\
LAB & 2.98 & 5.19 & 3.49 \\
DU-Net & 2.97 & 5.53 & 3.57 \\
DeCaFa & 2.93 & 5.26 & 3.39 \\
Teacher & 2.91 & 5.91 & 3.49 \\
HR-Net & 2.87 & 5.15 & 3.32 \\
HG-HSLE & 2.85 & 5.03 & 3.28 \\
DCFE(w/3D) & 2.76 & 5.22 & 3.24 \\
LUVli & 2.76 & 5.16 & 3.23 \\
AWing & 2.72 & 4.52 & 3.07 \\
ADNet & \underline{2.53} & 4.58 &  \underline{2.93} \\
\hline
Scratch & 2.90 & 5.19 & 3.35 \\
FaRL & 2.70 & 4.64 & 3.08 \\
Ours$_{0.1}$ & 2.65 & 4.68 & 3.06\\
Ours & 2.60 & 4.51 & 2.98 \\
\hline
FaRL$^{448}$ & 2.56 & \bf{4.45} & \underline{2.93} \\
Ours $_{0.1}^{448}$ & 2.57 & 4.66 & 2.98\\
Ours $^{448}$ & \bf {2.51} & \underline{4.47} & \bf{2.90}\\
\hline
\end{tabular}
\end{center}
\end{table}

Face alignment predicts 2D face landmark coordinates on a face image. Three datasets are used in our experiments: 
ALFW-19~\cite{alfw-19} with 20K images for training and 4,386 \pjl{images} for testing, \pjl{in which} each image is annotated with 19 landmarks;
WFLW~\cite{wlfw} with 7,500 images for training and 2,500 \pjl{images} for testing, \pjl{in which} each image is annotated with 98 landmarks;
300W~\cite{300w} with 3,837 training images and 600 testing images, \pjl{in which} each image is annotated with 68 landmarks.
We use a UperNet head to generate a heat map to predict the landmark position of the above face alignment tasks.
We use the \{2, 4, 8, 12\}th layer output of the ViT-b/16 to build the feature pyramid network, and during parse training we transfer the pre-trained model to fit the alignment loss.
We \pjl{apply the} normalized mean error (NME), failure rate (FR), and AUC to measure the performance.

For AFLW-19 dataset, we evaluate two kinds of face data:
(1) the full dataset;
and (2) the data that only contains the frontal face.
We report NME$_{diag}$, NME$_{box}$ and AUC$^7_{box}$ for alignment accuracy.
As shown in Table~\ref{tab:AFLW}, we can find  our model outperforms the FaRL pre-trained with the full LAION-FACE 20M dataset on $224\times224$ resolution.
Even we only used 10 percent of LAION-FACE-cropped during pre-train stage, we can get a model better than its counterpart FaRL on $224\times224$ resolution downstream task.
Our $448\times448$ model also surpasses FaRL (full NME$_{diag}$ 0.943 $\to$ 0.932, front NME$_{diag}$ 0.821 $\to$ 0.810, NME$_{box}$ 1.334 $\to$ 1.318) and get new state-of-the-art results.

WFLW dataset has 7 face categories, including 6 challenging categories and the full category.
We separately compare our method with previous SOTA methods on all of the categories and report the results in Table~\ref{tab:WFLW}.
We use NME$_{ioc}$ for single category performance evaluation and report FR$^{10}$ and AUC$^{10}$ as an extra metric for the full dataset.
Our model gets new state-of-the-art on all of the 7 categories of the face.
And our method surpasses FaRL with only 10\% LAION-fACE-cropped data  on $224\times224$ resolution downstream task.
Our $224\times224$ models outperform the $448\times448$ FaRL on 6 of 7 categories and get the same score on full categories.
For the same challenge task, we get significant improvement, e.g. Large Pose 6.91 $\to$ 6.55.

We also evaluate two kinds of face data for the iBUG-300 face challenges:
(1) the common face; 
(2) the challenge face.
We report NME$_{ioc}$ here to compare with previous work, and the results are shown in Table~\ref{tab:300W}.
Although the metric is getting saturated, we still get the best results on common and full categories.

\subsubsection{Face Parsing}

\begin{table}[t]
\begin{center}
\caption{Comparison with the state-of-the-art LaPa face parsing methods. Our method outperforms previous methods in most categories.}
\label{tab:LaPa}
\setlength\tabcolsep{1.5pt}
\renewcommand\arraystretch{1}
\footnotesize
% \scriptsize
\begin{tabular}{l|cccccccccc|c}
\hline
Method & Skin & Hair & L-E & R-E & U-L & I-M & L-L & Nose & L-B & R-B & Mean \\
\hline
EHANet & 95.8 & 94.3 & 87.0 & 89.1 & 85.3 & 85.6 & 88.8 & 94.3 & 85.9 & 86.1 & 89.2 \\
Wei et al. & 96.1 & 95.1 & 88.9 & 87.5 & 83.1 & 89.2 & 83.8 & 96.1 & 86.0 & 87.8 & 89.4 \\
BASS & 97.2 & 96.3 & 88.1 & 88.0 & 84.4 & 87.6 & 85.7 & 95.5 & 87.7 & 87.6 & 89.8 \\
EAGR & 97.3 & 96.2 & 89.5 & 90.0 & 88.1 & 90.0 & 89.0 & 97.1 & 86.5 & 87.0 & 91.1 \\
AGRNet & 97.7 & 96.5 & 91.6 & 91.1 & 88.5 & 90.7 & 90.1 & 97.3 & 89.9 & 90.0 & 92.3 \\
% FaRL* & 97.5 & 95.1 & 92.8 & 92.7 & 88.5 & 90.5 & {\bf 90.1} & 97.4 & 91.1 & 91.0 & 92.7 \\
\hline
Scratch  & 97.04 & 92.75 & 91.72 & 91.40 & 87.00 & 88.88 & 88.72 & 97.20 & 88.98 & 89.50 & 91.43 \\
FaRL  &  97.52 & 95.11 & 92.33 & 92.09 & 88.69 & 90.70 & 90.05 & 97.55 & 91.57 & 91.34 & 92.70  \\
Ours$_{0.1}$   & 97.46 & 94.93 & 92.26 & 92.22 & 88.31 & 90.20 & 89.84 & 97.53 & 91.20 & 90.79 &  92.48 \\
Ours  & 97.66 & 95.49 & 92.83 & 92.63 & 89.00 & 90.59 & 90.26 & 97.70 & 91.76 & 91.38 & 92.93 \\
\hline
FaRL$^{448}$  &\underline{98.00} & \underline{96.52} & \underline{93.97} & \underline{93.91} & \underline{90.15} & \underline{91.74} & \underline{91.21} & \underline{97.92} & \bf{92.70} & \bf{92.65} & \underline{93.88}  \\
Ours$^{448}_{0.1}$  & 97.91 & 96.36 & 93.53 & 93.42 & 90.01 & 91.47 & 90.89 & 97.85 & 92.34 & 91.74 & 93.55  \\
Ours$^{448}$  &  \bf{98.03} & \bf{96.69} & \bf{94.12} & \bf{93.95} & \bf{90.46} & \bf{91.98} & \bf{91.34} & \bf{97.95} & \underline{92.57} & \underline{92.44} & \bf{93.96} \\
\hline
\end{tabular}
\end{center}
% \vspace{-10pt}
\end{table}

\begin{table}[t]
\begin{center}
\caption{Comparison with the state-of-the-art CelebAMask-HQ face parsing methods. We get a SOTA mean F1 score.}
\label{tab:celebam}
\setlength\tabcolsep{1.5pt}
\renewcommand\arraystretch{1.2}
\footnotesize
\begin{tabular}{l|ccccccccc|c}
\hline
\multirow{2}*{Method} & Face & Nose & Glasses & L-Eye & R-Eye & L-B & R-B & L-Ear & R-Ear & \multirow{2}*{Mean} \\
~ & I-M & U-L & L-L & Hair & Hat & Earring & Necklace & Neck & Cloth & ~\\
\hline
\multirow{2}*{Zhao et al.} & 95.5 & 85.6 & 92.9 & 84.3 & 85.2 & 81.4 & 81.2 & 84.9 & 83.1 &  \multirow{2}*{80.3} \\
& 63.4 & 88.9 & 90.1 & 86.6 & 91.3 & 63.2 & 26.1 & 92.8 & 68.3 &  ~ \\
\hline
\multirow{2}*{Wei et al.} & 96.4 & 91.9 & 89.5 & 87.1 & 85.0 & 80.8 & 82.5 & 84.1 & 83.3 &  \multirow{2}*{82.1} \\
& 90.6 & 87.9 & 91.0 & 91.1 & 83.9 & 65.4 & 17.8 & 88.1 & 80.6 &  ~ \\
\hline
\multirow{2}*{EHANet} & 96.0 & 93.7 & 90.6 & 86.2 & 86.5 & 83.2 & 83.1 & 86.5 & 84.1 &  \multirow{2}*{84.0} \\
~& 93.8 & 88.6 & 90.3 & 93.9 & 85.9 & 67.8 & 30.1 & 88.8 & 83.5 &  ~ \\
\hline
\multirow{2}*{EAGR} & 96.2 & 94.0 & 92.3 & 88.6 & 88.7 & 85.7 & 85.2 & 88.0 & 85.7 &  \multirow{2}*{85.1} \\
~& 95.0 & 88.9 & 91.2 & 94.9 & 87.6 & 68.3 & 27.6 & 89.4 & 85.3 &  ~ \\
\hline
\multirow{2}*{AGRNet} & 96.5 & 93.9 & 91.8 & 88.7 & 89.1 & 85.5 & 85.6 & 88.1 & 88.7 &  \multirow{2}*{85.5} \\
~ & 92.0 & 89.1 & 91.1 & 95.2 & 87.2 & 69.6 & 32.8 & 89.9 & 84.9 &  ~ \\
\hline
\multirow{2}*{Scratch} & 96.17 & 93.77 & 92.28 & 89.04 & 88.97 & 85.32 & 85.36 & 86.88 & 87.32 &  \multirow{2}*{84.74}  \\
~ & 91.66 & 88.10 & 90.04 & 94.94 & 82.73 & 63.05 & 33.52 & 90.76 & 85.92 &  ~ \\
\hline
\multirow{2}*{FaRL} &  96.32 & 93.62 & 94.08 & 88.81 & 88.67 & 85.25 & 85.46 & 87.53 & 87.87 & \multirow{2}*{87.55}  \\
~ & 91.10 & 87.77 & 89.81 & 95.76 & 90.80 & 69.87 & 60.91 & 91.79 & 90.40 & ~ \\
\hline
\multirow{2}*{Ours$_{0.1}$} & 96.38 & 93.86 & 93.43 & 89.00 & 88.90 & 85.58 & 85.73 & 87.47  & 87.98 &  \multirow{2}*{86.69} \\
~ & 91.33 & 87.98 & 90.12 & 95.69 & 89.91 & 67.28 & 48.68 & 91.65 & 89.44 &  ~ \\
\hline
\multirow{2}*{Ours} & 96.51 & 93.92 & 94.55 & 89.54 & 89.48  & 85.91 & 86.09 & 88.01 & 88.38 & \multirow{2}*{87.91}  \\
~ & 92.08 & 88.89 & 90.70 & 95.88 & 89.90 & 71.91 & 58.12 & \bf{92.70} & 90.41 &  ~ \\
\hline
\multirow{2}*{FaRL$^{448}$} & \underline{96.74} & 94.22 & \underline{95.37} & 90.71 & 90.56 & 87.03 & \underline{87.14} & 89.06 & \underline{89.46} & \multirow{2}*{\underline{89.56}}   \\
~ & 92.80 & 90.17 & 91.38 & \underline{96.20} & \bf{92.09} & \underline{75.72} & \underline{69.72} & \underline{92.45} & \bf{91.31} &  ~ \\
\hline
\multirow{2}*{ Ours$_{0.1}^{448}$} & \underline{96.74} & \underline{94.25} & 95.17 & \underline{90.74} & \underline{90.59} & 87.00 & 87.00 & \underline{89.07} & 89.35 & \multirow{2}*{88.83}  \\
~ & \underline{92.95} & \underline{90.23} & \underline{91.51} & 96.06 & 90.19 & 74.43 & 61.04 & 92.36 & 90.38 &  ~ \\
\hline
\multirow{2}*{Ours$^{448}$} & \bf{96.83} & \bf{94.34} & \bf{95.48} & \bf{90.91} & \bf{90.76} & \bf{87.23} & \bf{87.35} & \bf{89.37} & \bf{89.55} &  \multirow{2}*{\bf{89.77}} \\
~ & \bf{93.16} & \bf{90.49} & \bf{91.67} & \bf{96.26} & \underline{91.59} & \bf{76.70} & \bf{70.43} & \underline{92.55} & \underline{91.25} &  ~ \\
\hline
\end{tabular}
% \vspace{-15pt}
\end{center}
\end{table}

Face parsing segments face into different regions which correspond to different components. 
We test our model on two face parsing datasets: LaPa~\cite{lapa} and CelebAMask-HQ~\cite{celebamask}.
Lapa contains 18,176 training images and 2K test images, and each face image is annotated with 11 face regions.
CelebAmask-HQ contains 24,183 training images in $1024\times1024$ resolution and 2,824 \pjl{testing images}. Each face is divided into 19 regions.
We follow the setting of FaRL~\cite{FaRL} to construct the parse network, which uses UperNet~\cite{upernet} for segmentation.
And we construct feature pyramid network as the same as the face alignment experiment and transfer the pre-trained backbone to a certain task.
We report the F1 score of facial components to measure the model performance.

For LaPa dataset, we report the parsing accuracy of all the 10 face regions with F1 score, and compare our results with previous supervised methods and recent lingual-visual supervised method FaRL~\cite{FaRL}.
We report the results under both light setting and full setting pre-train.
For a fair comparison, we also report the results with $448\times448$ resolution during transfer training.
As shown in Table~\ref{tab:LaPa}, we can find that our light pre-trained model outperforms the previous supervised method in both $224\times224$ and $448\times448$ resolution.
Compared with the scratch model, the light pre-trained model has about 0.9\% improvement in mean F1 score and the full pre-trained model has 1.3\% improvement.
Our full pre-trained model outperforms the results of FaRL, and for $224\times224$ resolution, we get 0.23 mean F1 improvement (92.70 $\to$ 92.93), and 0.08 improvement (93.88 $\to$ 93.96) under $448\times448$ resolution.

For CelebAMask-HQ dataset, we follow the same evaluation setting as LaPa. Compared with LaPa dataset, CelebAMask-HQ has more categories for face parsing. Experiments show that our model has the best performance on this dataset.
As shown in Table~\ref{tab:celebam}, our method outperforms the state-of-the-art results in most categories and gets the best mean F1 score (89.56 $\to$ 89.77).
We find that as the pre-trained dataset gets larger, the segmentation accuracy of the Necklace region gets more significant improvement.

\subsection{Comparison with General Representation Learning}

Our Mask Contrastive Face (MCF) is designed for face domain representation learning,
so we focus on comparing our method with previous related SOTA face domain methods, \textit{i.e.}, FaRL~\cite{FaRL}. 
And we have compared our method with some general methods, \textit{i.e.},  CLIP, in Table.\ref{tab:compare_general}.
For fair comparison, we pre-train the backbone with MAE and 10\% of our dataset, 
which got 91.65 LaPa F1(0.83$\downarrow$ worse than ours) and 1.031 AFLW-19 NME$_{diag}$ (0.64 $\uparrow$ worse than ours). Previous research~\cite{FaRL} has given the comparison results with other representation methods (freezing backbone), 
and we illustrate part of the results in Table~\ref{tab:compare_general},
where for LaPa F1-mean, we got 92.74 (0.88 $\uparrow$ than MoCo v3) 
and for AFLW-19 NME$_{diag}$, we got 0.954 (0.041 $\downarrow$ than SimCLR).
The comparison results demonstrate that our method is more suitable for face domain representation learning.

\begin{table}[t]
\begin{center}
\caption{Comparison with previous representation learning. Marked * results come from ~\cite{FaRL}. Our method outperforms all previous methods. (both general and face).}
\label{tab:compare_general}
\footnotesize
% \scriptsize
% \small
\begin{tabular}{l|c|cc}
\hline
Method & Pre-train data & LaPa F1 $\uparrow$ & AFLW-19 NME$_{diag}$ $\downarrow$\\
\hline
MoCo v3$^*_{Freeze}$ & general & 91.86 & 1.007\\
DeiT$^*_{Freeze}$ & general & 92.00 & 1.003 \\ 
CLIP$^*_{Freeze}$ & general & 92.21 & 0.995 \\ 
SimCLR$^*_{Freeze}$ & face & 91.72 & 0.995 \\ 
Face Transformer$^*_{Freeze}$ & face & 91.09 & 1.031 \\ 
FaRL$^*_{Freeze}$ & face & 92.32 & 0.991 \\
Ours$_{Freeze}$ & face & \bf 92.74 & \bf 0.954 \\
\hline
MAE$_{0.1}$ & face &  91.65 & 1.031 \\
Ours$_{0.1}$ & face & \bf 92.48 & \bf 0.967 \\
\hline
\end{tabular}
\end{center}
\end{table}
\subsection{Ablation Study}
\label{sec:ablation}

To verify the effectiveness of each module, we conduct several ablation studies on the key technical contributions.
(1) $\mathcal{L}_2$, a mask image modeling with $L_2$ loss, which is close to MAE~\cite{mae}, and in order to learn the globe face identity, we remove the random resize crop operation during data augmentation;
(2) $\mathcal{L}_2$ + $\mathcal{L}_{mim}$, which directly adds the $\mathcal{M}_1$ decoder at the end of MAE decoder;
(3) $\mathcal{L}_{mim}$ , a mask image model with 8-layer $\mathcal{M}_0$ and a $\mathcal{M}_1$ decoder;
(4) $\mathcal{L}_{mim}$ , a mask image model with 2-layer $\mathcal{M}_0$ and a $\mathcal{M}_1$ decoder;
(5) $\mathcal{L}_{mim}$ + $Con_{p}$ (Ours), a framework contains a mask image modeling with $\mathcal{M}_0^2$ and a contrastive strategy on patch tokens;
(6) $\mathcal{L}_{mim}$ + $Con_{cls}$ (Ours), a framework contains a mask image modeling with $\mathcal{M}_0^2$ and a contrastive strategy on class tokens.

The ablation study results are illustrated in Table~\ref{tab:ablation}, where all of the ablation studies are pre-trained on the light LAION-FACE-cropped dataset.
Comparing row 2 and row 3, we can find that mixed mask image modeling has a negative influence on downstream facial analysis tasks.
And row 4 shows that our partial learnable decoder plays an important role in promoting the model performance, and has a 0.57\% improvement (91.65 $\to$ 92.22) on LaPa face parsing task.
Decreasing the decoder depth to our final setting gets another 0.17\% improvement.
Finally, we verify the contrastive strategy, and find that our class token based contrastive strategy achieves higher performance on downstream tasks than the patch token based one.
Our full setting results are illustrated on the last row of Table~\ref{tab:ablation}, which has 1\% improvement than the scratch model on LaPa face parsing task, and has about 0.1\% improvement (92.39 $\to$ 92.48) than the model without contrastive learning.  

\begin{table}[t]
\begin{center}
\caption{Ablation study of key technical contributions. We evaluate the model performance on LaPa and AFLW-19 tasks. Both mask image modeling and contrastive strategy play important roles in promoting downstream task performance.}
\label{tab:ablation}
\setlength\tabcolsep{1pt}
\renewcommand\arraystretch{1}
% \scriptsize
\footnotesize
\begin{tabular}{l|c|c|cccc}
\hline
\multirow{2}*{Method} & {$\mathcal{M}_0$} & \multicolumn{1}{c|}{Lapa} & \multicolumn{4}{c}{AFLW19} \\
~ & depth & F1-mean$\uparrow$ & NME$_{diag}^{full}$ $\downarrow$ & NME$_{diag}^{front}$ $\downarrow$ & NME$_{box}$  $\downarrow$ & AUC$^{7}_{box}$ $\uparrow$ \\
\hline
Scratch & 8 & 91.43 & 1.047 & 0.884 & 1.481 & 79.3  \\
$\mathcal{L}_2$  & 8 & 91.65 & 1.031 & 0.873 & 1.459 & 79.6\\
$\mathcal{L}_2$ + $\mathcal{L}_{mim}$ & 8 & 91.55 & 1.021 & 0.868 & 1.447 & 79.8\\
$\mathcal{L}_{mim}$ & 8 & 92.22 & 0.998 & 0.848 & 1.411 & 80.2 \\
\hline
$\mathcal{L}_{mim}$ & 2 & 92.39 & 0.972 & 0.831 & 1.374 & 80.8\\
$\mathcal{L}_{mim}$ + Con$_{p}$  & 2 & 92.41 & 0.968 & 0.830 & \bf 1.368 & \bf 80.8\\
$\mathcal{L}_{mim}$ + Con$_{cls}$  & 2 & \bf 92.48 & \bf 0.967 & \bf 0.827 & \bf 1.368 & \bf 80.8 \\
\hline
\end{tabular}
\end{center}
% \vspace{-5pt}
\end{table}

\begin{figure}[t]
    \includegraphics[width=0.85\columnwidth]{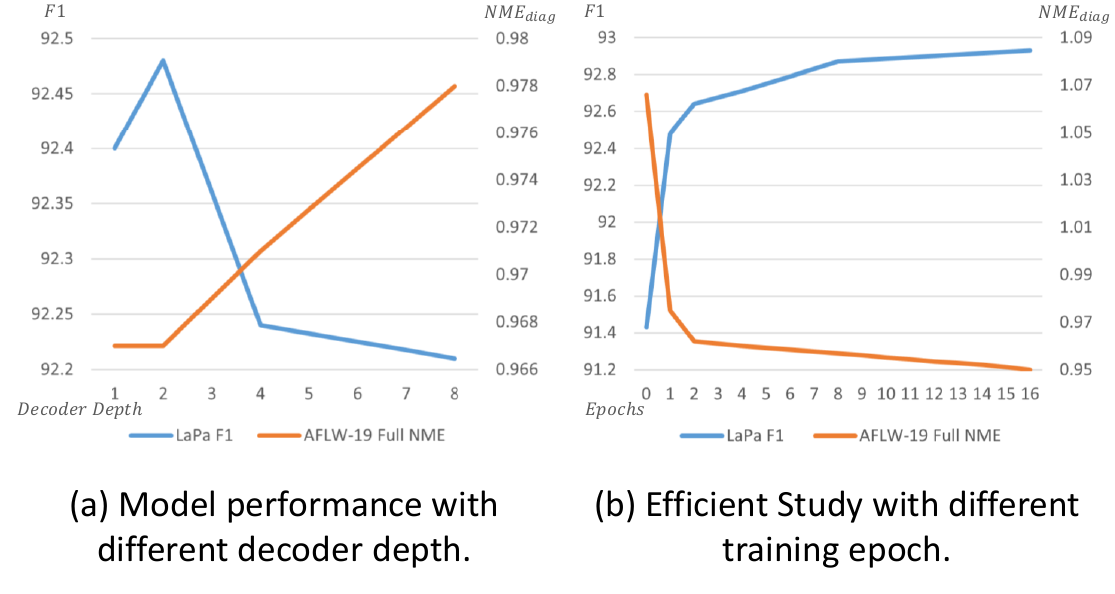}
    % \vspace{-20pt}    
    \caption{Performance trend of different decoder depth and training epoch. The $y$ axis in left figure is Lapa F1 score and the $y$ axis in right figure is AFLW-19 NME$_{diag}^{full}$ score.}
    \label{fig:analysis}
\end{figure}

\subsection{Learnable Decoder Depth}

Decoder depth determined how many degrees of freedom in Eq.(\ref{eq:linear}), we explore the influence of learnable decoder depth in this section.
We conduct experiments using 4 different $M_0$ decoders separately with \{1, 2, 4, 8\} layers and pre-train the model on 10\% of LAION-FACE-cropped dataset.
We evaluate the model performance on LaPa face parsing and AFLW-19 face alignment tasks. 
As shown in %Table~\ref{tab:Learnable_Decoder_Depth} and 
Figure~\ref{fig:analysis}(a), 
the 2-layer decoder achieves the best performance.
And as the decoder $\mathcal{M}_0$ gets deeper from 2 layers, the performance decreases gradually.
We think the reason is that as the degree of freedom increases, the optimization becomes further from the linear programming task, and the influence of $\mathcal{M}_1$ gets weaker.

\subsection{Efficiency Study}

In this section, we explore the training efficiency of our framework. In detail, we evaluate different model training epochs with two downstream face analysis tasks as the same as Section~\ref{sec:ablation}.
The result is illustrate as %Table~\ref{tab:Efficiency_Study} and 
Figure~\ref{fig:analysis}(b).
We pre-train our model on the full LAION-FACE-cropped dataset.
We can find that when pre-trained only 2 epochs, our model outperforms FaRL on AFLW-19 datasets; when pre-trained 4 epochs, our model outperforms FaRL on LaPa datasets, which means our method has a better data efficiency during training.
As the number of training epoch increases, the model performance improvement becomes smaller.

\subsection{Results with ImageNet Pre-train}
\label{sec:in1k}

In this section, we explore the performance of the model pre-trained with both ImageNet 1K and LAION-FACE-cropped datasets.
In detail, we use a pre-trained ViT-b/16 (with MAE~\cite{mae} on ImageNet 1K for 1600 epochs) as initial parameters and refine on LAION-FACE-cropped for 16 epochs with our Mask Contrastive Face.
The comparison results are illustrated in Tables~\ref{tab:AFLW-in1k}~\ref{tab:celebam-in1k}, more results will be illustrated in appendix.
Ours$_{+IN1K}$ indicates the ImageNet 1K pre-trained version. 
Ours$_{+IN1K\&Freeze}$ means freezing the backbone during downstream training.
Ours$^{448}_{IN1K}$ means 448 resolution version downstream task.

With pre-training on ImageNet 1K but not freezing the backbone parameters during downstream tasks (Ours$_{+IN1K}$), compared with only pre-trained on LAION-FACE-cropped, some downstream tasks have improved results but the others have decreased results (e.g., NME$_{diag}$ 448 version in Table~\ref{tab:AFLW-in1k}, mean F1 score 448 version in Table~\ref{tab:celebam-in1k}).
Considering that ImageNet pre-training costs much more computation resources than the Mask Contrastive Face stage, we can assume that the role of ImageNet pre-training is limited.
We also find that if we freeze the backbone parameters during downstream tasks (Ours$_{IN1K\&Freeze}$), we can still get better results compared with previous methods (e.g. -0.015 NME$_{diag}$ on AFLW-19 better than FaRL and +0.19 mean F1 Score on CelebAMask-HQ).% which means that our pre-trained model outputs a great face representation.% without any refining. 

\begin{table}[t]
\begin{center}
\footnotesize
\caption{Results of different pre-training and fine-tuning settings on AFLW-19 face alignment methods. Firstly pre-training with ImageNet-1K has no significant influence.}
\begin{tabular}{l|cc|cc}
\hline
\multirow{2}*{Method} & \multicolumn{2}{c|}{NME$_{diag}$ $\downarrow$}  & NME$_{box}$  $\downarrow$ & AUC$^{7}_{box}$ $\uparrow$\\
~ & Full & Front & Full & Full\\
\hline
Scratch & 1.047 & 0.884 & 1.481 & 79.3 \\
FaRL & 0.969 & 0.836 & 1.371 & 80.8 \\
Ours $_{0.1}$ & 0.968 & 0.827 & 1.368 & 80.8 \\
Ours & 0.950 & 0.820 & 1.344 & 81.2 \\
Ours$_{+IN1K\&Freeze}$ & 0.954 & 0.832 & 1.349 & 81.1 \\
Ours$_{+IN1K}$ & 0.949 & 0.827 & 1.342 & 81.2 \\
\hline
FaRL$^{448}$ & 0.943 & 0.821 & 1.334 & 81.3\\
Ours$_{0.1}^{448}$ & 0.951 & 0.825 & 1.345 & 81.2 \\
Ours$^{448}$ & 0.932 & 0.810 & 1.318 &  81.5\\
Ours$^{448}_{+IN1K}$ & 0.933 & 0.810 & 1.320 &  81.6\\
% Ours & {\bf 0.975} & {\bf 0.831} & {\bf 1.38} & {\bf 80.7}\\
\hline
\end{tabular}
% \vspace{-10pt}
\label{tab:AFLW-in1k}
\end{center}
\end{table}

\begin{table}[t]
\begin{center}
\caption{Results of different pre-training and finetune settings on CelebAMask-HQ face parsing methods.  Firstly pre-training with ImageNet-1K has no significant influence.}
\setlength\tabcolsep{1.5pt}
\renewcommand\arraystretch{1}
\scriptsize
\begin{tabular}{l|ccccccccc|c}
\hline
\multirow{2}*{Method} & Face & Nose & Glasses & L-Eye & R-Eye & L-B & R-B & L-Ear & R-Ear & \multirow{2}*{Mean} \\
~ & I-M & U-L & L-L & Hair & Hat & Earring & Necklace & Neck & Cloth & ~\\
\hline
% \multirow{2}*{FaRL*} & 96.5 & {\bf 94.0} & 94.1 & 90.0 & {\bf 90.0} & 86.3 & 86.2 & 88.0 & 88.4 &  \multirow{2}*{87.4} \\
% ~& 92.1 & 89.4 & 90.8 & 95.6 & 87.4 & 69.2 & {\bf 55.0} & 91.8 & 89.1 &  ~ \\
% \hline
% \multirow{2}*{Ours} & {\bf 96.6} &{\bf 94.0} & {\bf 94.7} & {\bf 90.1} & {\bf 90.0} & {\bf 86.4} & {\bf 86.5} & {\bf 88.4} & {\bf 88.8} &  \multirow{2}*{{\bf 87.7}} \\
% & {\bf 92.4} & {\bf 89.6} & {\bf 91.0} & {\bf 95.8} & {\bf 88.4} & {\bf 71.6} & \underline{51.8} & {\bf 92.0} & {\bf 90.0} &  ~ \\ 
\multirow{2}*{Scratch} & 96.17 & 93.77 & 92.28 & 89.04 & 88.97 & 85.32 & 85.36 & 86.88 & 87.32 &  \multirow{2}*{84.74}  \\
~ & 91.66 & 88.10 & 90.04 & 94.94 & 82.73 & 63.05 & 33.52 & 90.76 & 85.92 &  ~ \\
\hline
\multirow{2}*{FaRL} &  96.32 & 93.62 & 94.08 & 88.81 & 88.67 & 85.25 & 85.46 & 87.53 & 87.87 & \multirow{2}*{87.55}  \\
~ & 91.10 & 87.77 & 89.81 & 95.76 & 90.80 & 69.87 & 60.91 & 91.79 & 90.40 & ~ \\
\hline
\multirow{2}*{Ours$_{0.1}$} & 96.38 & 93.86 & 93.43 & 89.00 & 88.90 & 85.58 & 85.73 & 87.47  & 87.98 &  \multirow{2}*{86.69} \\
~ & 91.33 & 87.98 & 90.12 & 95.69 & 89.91 & 67.28 & 48.68 & 91.65 & 89.44 &  ~ \\
\hline
\multirow{2}*{Ours} & 96.51 & 93.92 & 94.55 & 89.54 & 89.48  & 85.91 & 86.09 & 88.01 & 88.38 & \multirow{2}*{87.91}  \\
~ & 92.08 & 88.89 & 90.70 & 95.88 & 89.90 & 71.91 & 58.12 & 92.70 & 90.41 &  ~ \\
\hline
\multirow{2}*{Ours$_{+IN1K\&Freeze}$} & 96.54 & 94.05 & 94.38 & 89.53 & 89.39  & 86.11 & 86.21 & 88.20 & 88.56 & \multirow{2}*{87.74}  \\
~ & 92.02 & 88.92 & 90.74 & 95.83 & 89.07 & 70.73 & 56.44 & 92.09 & 90.44 &  ~ \\
\hline
\multirow{2}*{Ours$_{+IN1K}$} & 96.55 & 94.03 & 94.62 & 89.59 & 89.45  & 86.14 & 86.45 & 88.32 & 88.52 & \multirow{2}*{88.17}  \\
~ & 92.11 & 88.98 & 90.75 & 95.98 &90.69 & 72.26 & 59.90 & 92.12 & 90.83 &  ~ \\
\hline
\multirow{2}*{FaRL$^{448}$} & {96.74} & 94.22 & {95.37} & 90.71 & 90.56 & 87.03 & {87.14} & 89.06 & {89.46} & \multirow{2}*{{89.56}}   \\
~ & 92.80 & 90.17 & 91.38 & 96.20 & 92.09 & {75.72} & 69.72 & {92.45} & {91.31} &  ~ \\
\hline
\multirow{2}*{ Ours$_{0.1}^{448}$} & {96.74} & {94.25} & 95.17 & {90.74} & {90.59} & 87.00 & 87.00 & {89.07} & 89.35 & \multirow{2}*{88.83}  \\
~ & {92.95} & {90.23} & {91.51} & 96.06 & 90.19 & 74.43 & 61.04 & 92.36 & 90.38 &  ~ \\
\hline
\multirow{2}*{Ours$^{448}$} & {96.83} & {94.34} & {95.48} & {90.91} & {90.76} & 87.23 & 87.35 & 89.37 & {89.55} &  \multirow{2}*{{89.77}} \\
~ & 93.16 & {90.49} & {91.67} & {96.26} & {91.59} & {76.70} & {70.43} & {92.55} & {91.25} &  ~ \\
\hline
\multirow{2}*{Ours$^{448}_{+IN1K}$} & 96.82 & 94.31 & 95.60 & 90.98 & 90.85 & 87.33 & 87.40 & 89.15 & 89.54 & \multirow{2}*{89.68}  \\
~ & 93.23 & 90.52 & 91.72 & 96.27 & 91.41 & 76.63 & 68.83 & {92.63} & 90.99 &  ~ \\
\hline
\end{tabular}
% \vspace{-10pt}

\label{tab:celebam-in1k}
\end{center}
\end{table}

\section{Conclusion}
In this paper, we propose a face self-supervised representation learning framework called Mask Contrastive Face (MCF), which combines improved mask image modeling and contrastive learning. 
For better facial representation learning research, we construct a fixed and face-aligned variant of the LAION-FACE 20M dataset called the LAION-FACE-cropped dataset.
We pre-train the visual backbone with Mask Contrastive Face (MCF) on the proposed dataset and validate its performance on multiple downstream tasks. 
% We validated its performance on downstream tasks and showed that it outperforms previous methods. 
We hope that our work can further improve the accuracy of face analysis tasks and contribute to the development of facial analysis research.

\begin{acks}
This work is supported by National Natural Science Foundation of China (72192821, 62272447), Shanghai Sailing Program (22YF1420300), Shanghai Municipal Science and Technology Major Project (2021SHZ\\DZX0102), CCF-Tencent Open Research Fund (RAGR20220121), Young Elite Scientists Sponsorship Program by CAST (2022QNRC001), Beijing Natural Science Foundation (L222117), the Fundamental Research Funds for the Central Universities (YG2023QNB17).
\end{acks}

\bibliographystyle{ACM-Reference-Format}
\bibliography{sample-base}

\clearpage
\appendix

\section{More Analysis Results}
\label{sec:eff}
In this section, we give more analysis results.
In Table~\ref{tab:Learnable_Decoder_Depth}, we show influence of learnable decoder $\mathcal{M}_0$ depth. We get the best results with a 2-layer decoder on both downstream tasks.
In Table~\ref{tab:Efficiency_Study}, we give the results of evaluating different model training epochs with two downstream face analysis tasks as the same as Section 4.5.

\section{More Results with ImageNet Pre-train}
\label{sec:in1k}

In this section, we explore the performance of the model pre-trained with both ImageNet 1K and LAION-FACE-cropped datasets.
In detail, we use a pre-trained ViT-b/16 (with MAE~\cite{mae} on ImageNet 1K for 1600 epochs) as initial parameters and refine on LAION-FACE-cropped for 16 epochs with our Mask Contrastive Face.
The comparison results are illustrated in Tables~\ref{tab:AFLW-in1k}~\ref{tab:300W-in1k}~\ref{tab:WFLW-in1k}~\ref{tab:lapa-in1k}~\ref{tab:celebam-in1k}.
Ours$_{+IN1K}$ indicates the ImageNet 1K pre-trained version. 
Ours$_{+IN1K\&Freeze}$ means freezing the backbone during downstream training.
Ours$^{448}_{IN1K}$ means 448 resolution version downstream task.

With pre-training on ImageNet 1K but not freezing the backbone parameters during downstream tasks (Ours$_{+IN1K}$), compared with only pre-trained on LAION-FACE-cropped, some downstream tasks have improved results but the others have decreased results (e.g., NME$_{diag}$ 448 version in Table~\ref{tab:AFLW-in1k}, mean F1 score 448 version in Table~\ref{tab:celebam-in1k}).
Considering that ImageNet pre-training costs much more computation resources than the Mask Contrastive Face stage, we can assume that the role of ImageNet pre-training is limited.

We find that if we freeze the backbone parameters during downstream tasks (Ours$_{+IN1K\&Freeze}$), we can still get better results compared with previous methods (e.g. FaRL), which means that our pre-trained model outputs a great face representation without any refining. 

\begin{table}[t]
\begin{center}
\caption{Influence of learnable decoder $\mathcal{M}_0$ depth. We get the best results with a 2-layer decoder on both downstream tasks.}
\label{tab:Learnable_Decoder_Depth}
\setlength\tabcolsep{3pt}
\renewcommand\arraystretch{1}
\footnotesize
\begin{tabular}{l|c|cccc}
\hline
\multirow{2}*{Depth} & \multicolumn{1}{c|}{Lapa} & \multicolumn{4}{c}{AFLW19} \\
~ & F1-mean$\uparrow$ & NME$_{diag}^{full}$ $\downarrow$ & NME$_{diag}^{front}$ $\downarrow$ & NME$_{box}$  $\downarrow$ & AUC$^{7}_{box}$ $\uparrow$ \\
\hline
1 & 92.4 & 0.967 & 0.825 & 1.368 & 80.8 \\
2 & 92.48 & 0.967 & 0.827 & 1.368 & 80.8 \\
4 & 92.24 & 0.971 & 0.834 & 1.373 & 80.7 \\
8 & 92.21 & 0.978 & 0.837 & 1.384 & 80.6 \\
\hline
\end{tabular}
\end{center}
\end{table}

\begin{table}[t]
\begin{center}
\caption{Training efficiency study with different train epochs. The model performance gets convergence when per-trained with 16 epochs.}
\label{tab:Efficiency_Study}
\setlength\tabcolsep{3pt}
\renewcommand\arraystretch{1}
% \scriptsize
\footnotesize
% \small
\begin{tabular}{l|c|cccc}
\hline
\multirow{2}*{Epochs} & \multicolumn{1}{c|}{Lapa} & \multicolumn{4}{c}{AFLW19} \\
~ & F1-mean$\uparrow$ & NME$_{diag}^{full}$ $\downarrow$ & NME$_{diag}^{front}$ $\downarrow$ & NME$_{box}$  $\downarrow$ & AUC$^{7}_{box}$ $\uparrow$ \\
\hline
Scratch & 91.43 & 1.047 & 0.884 & 1.481 & 79.3  \\
1 & 92.48 & 0.975 & 0.838 & 1.380 & 80.7  \\
2 & 92.64 & 0.962 & 0.827 & 1.361 & 80.9 \\
4 & 92.71 & 0.960 & 0.831 & 1.358 & 81.0\\
8 & 92.87 & 0.957 & 0.825 & 1.354 & 81.0\\
16 & 92.93 & 0.950 & 0.820 & 1.344 & 81.2  \\
\hline
\end{tabular}
\end{center}
\end{table}

\begin{table*}[t]
\begin{center}
\footnotesize
\caption{Comparison with the state-of-the-art iBUG-300 face alignment methods. Our method outperforms the previous methods on both common and full categories.}
\begin{tabular}{l|cc|c}
\hline
\multirow{2}*{Method} & \multicolumn{3}{c}{NME$_{ioc}$ $\downarrow$}\\
~ & Common & Challenge & Full\\
\hline
Scratch & 2.90 & 5.19 & 3.35 \\
FaRL & 2.70 & 4.64 & 3.08 \\
Ours$_{0.1}$ & 2.65 & 4.68 & 3.06\\
Ours & 2.60 & 4.51 & 2.98 \\
Ours$_{+IN1K\&Freeze}$ & 2.59 & 4.51 & 2.96 \\
Ours$_{+IN1K}$ & 2.60 & 4.55 & 2.98 \\
\hline
FaRL$^{448}$ & 2.56 & 4.45 & 2.93 \\
Ours$_{0.1}^{448}$ & 2.57 & 4.66 & 2.98\\
Ours$^{448}$ & 2.51 & 4.47 & 2.90\\
Ours$^{448}_{+IN1K}$ & 2.50 & 4.46 & 2.89 \\
\hline
\end{tabular}
\label{tab:300W-in1k}
\end{center}
\end{table*}

\begin{table*}[h]
\begin{center}
\footnotesize
\caption{Comparison with the state-of-the-art WFLW face alignment methods. Our method outperforms previous methods in both 224$\times$224 and 448$\times$448 resolution.}
\begin{tabular}{l|c|cccccc|cc}
\hline
\multirow{2}*{Method} & \multicolumn{7}{c|}{NME$_{ioc}$ $\downarrow$} &FR$^{10}\downarrow$ & AUC$^{10}\uparrow$ \\
~ & Full & Pose & Expr. & Illum. & MakeUp & Occl. & Blur & \multicolumn{2}{c}{Full}\\
\hline
% FaRL* & 4.39 & 7.78 & 4.65 & 4.31 & {\underline{4.35}} & 5.53 & 4.99 & 3.23 & 57.63 \\
Scratch & 4.80 & 8.78 & 5.09 & 4.74 & 4.99 & 6.01 & 5.35 & 5.72 & 54.54 \\
FaRL & 4.03 & 6.81 & 4.32 & 3.92 & 3.87 & 4.70 & 4.54 & 1.76 & 60.23 \\
Ours$_{0.1}$ & 4.16 & 7.15 & 4.40 & 4.05 & 4.09 & 4.99 & 4.66 & 2.56 & 59.19 \\
Ours & 3.96 & 6.61 & 4.13 & 3.84 & 3.80 & 4.68 & 4.43 & 1.40 & 60.90 \\
Ours$_{+IN1K\&Freeze}$ & 4.03 & 6.74 & 4.24 & 3.87 & 3.83 & 4.84 & 4.55 & 1.89 & 60.26 \\
Ours$_{+IN1K}$ & 3.94 & 6.54 & 4.12 & 3.81 & 3.75 & 4.62 & 4.44 & 1.44 & 60.94 \\
\hline
FaRL$^{448}$ & 3.96 & 6.91 & 4.21 & 3.97 & 3.80 & 4.71 & 4.57 & 1.76 &  61.16 \\
Ours$_{0.1}^{448}$ & 4.13 & 7.18 & 4.31 & 4.09 & 4.07 &  5.00 & 4.72 & 2.44 & 60.18  \\
Ours$^{448}$ & 3.90 & 6.55 & 4.07 & 3.82 & 3.72 & 4.61 & 4.42 & 1.60 & 61.44 \\
Ours$^{448}_{+IN1K}$ & 3.88 & 6.50 & 4.06 & 3.79 & 3.64 & 4.57 & 4.40 & 1.36 & 61.57 \\
\hline
\end{tabular}
\label{tab:WFLW-in1k}
\end{center}
\end{table*}

\begin{table*}[h]
\begin{center}
\footnotesize
\caption{Comparison with the state-of-the-art LaPa face parsing methods. Our method outperforms previous methods in most categories.}
\begin{tabular}{l|cccccccccc|c}
\hline
Method & Skin & Hair & L-E & R-E & U-L & I-M & L-L & Nose & L-B & R-B & Mean \\
\hline
Scratch  & 97.18 & 93.06 & 91.61 & 91.50 & 87.22 & 89.44 & 89.13 & 97.26 & 90.12 & 89.69 & 91.62 \\
FaRL  &  97.52 & 95.11 & 92.33 & 92.09 & 88.69 & 90.70 & 90.05 & 97.55 & 91.57 & 91.34 & 92.70  \\
Ours$_{0.1}$   & 97.46 & 94.93 & 92.26 & 92.22 & 88.31 & 90.20 & 89.84 & 97.53 & 91.20 & 90.79 &  92.48 \\
Ours  & 97.66 & 95.49 & 92.83 & 92.63 & 89.00 & 90.59 & 90.26 & 97.70 & 91.76 & 91.38 & 92.93 \\
Ours$_{+IN1K\&Freeze}$ & 97.59 & 95.38 & 92.24 & 92.24 & 89.05 & 91.24 & 90.44 & 97.61 & 90.71 & 90.94 & 92.74 \\
Ours$_{+IN1K}$ & 97.72 & 95.84 & 92.86 & 92.65 & 89.32 & 91.02 & 90.44 & 97.75 & 91.88 & 91.43 & 93.09 \\
\hline
FaRL$^{448}$  &98.00 & 96.52 & 93.97 & 93.91 & 90.15 & 91.74 & 91.21 & 97.92 & 92.70 & 92.65 & 93.88 \\
Ours$^{448}_{0.1}$  & 97.91 & 96.36 & 93.53 & 93.42 & 90.01 & 91.47 & 90.89 & 97.85 & 92.34 & 91.74 & 93.55  \\
Ours$^{448}$  &  98.03 & 96.69 & 94.12 & 93.95 & 90.46 & 91.98 & 91.34 & 97.95 & 92.57 & 92.44 & 93.96 \\
Ours$^{448}_{+IN1K}$ & 98.05 & 96.67 & 94.01 & 93.99 & 90.38 & 92.10 & 91.48 & 97.97 & 92.72 & 92.53 & 93.99 \\
\hline
\end{tabular}
\label{tab:lapa-in1k}
\end{center}
\end{table*}

\end{document}